\documentclass[11pt]{article}

\usepackage[final]{acl}

\usepackage{times}
\usepackage{latexsym}
\usepackage{float}

\usepackage[T1]{fontenc}

\usepackage[utf8]{inputenc}

\usepackage{microtype}

\usepackage{inconsolata}
\usepackage{amsmath}
\usepackage{booktabs}
\usepackage{placeins}

\usepackage{graphicx}
\usepackage{tikz}
\usetikzlibrary{shapes.geometric, arrows, positioning, arrows.meta}

\usepackage[dvipsnames]{xcolor}

\tikzstyle{startstop} = [rectangle, rounded corners, text width=1.8cm, align=center, minimum height=0.7cm, draw=black, fill=red!20]
\tikzstyle{process} = [rectangle, rounded corners, text width=1.8cm, align=center, minimum height=0.7cm, draw=black, fill=violet!20]
\tikzstyle{processgreen} = [rectangle, rounded corners, text width=1.8cm, align=center, minimum height=0.7cm, draw=black, fill=green!20]
\tikzstyle{processorange} = [rectangle, rounded corners, text width=1.8cm, align=center, minimum height=0.7cm, draw=black, fill=orange!30]
\tikzstyle{processyellow} = [rectangle, rounded corners, text width=1.8cm, align=center, minimum height=0.7cm, draw=black, fill=yellow!30]
\tikzstyle{processblue} = [rectangle, rounded corners, text width=1.8cm, align=center, minimum height=0.7cm, draw=black, fill=blue!20]
\tikzstyle{arrow} = [thick, -{Stealth}]

\title{BhashaSetu: A Data-Centric Approach to Low-Resource Machine Translation}

\author{
  Param Thakkar$^1$, Anushka Yadav$^1$, Michael Tiemann$^2$, Abhi Mehta$^1$, Akshita Bhasin$^1$, Shrinivas Khedkar$^1$ \\
  $^1$ Department of Computer Engineering and Information Technology,
  Veermata Jijabai Technological Institute, Mumbai \\
  $^2$ Tübingen AI Center, University of Tübingen, Germany \\
  \texttt{puthakkar\_b22@ce.vjti.ac.in}, \texttt{asyadav\_b22@ce.vjti.ac.in}, \texttt{michael.tiemann@uni-tuebingen.de} \\
  \texttt{armehta\_b22@it.vjti.ac.in}, \texttt{ambhasin\_b22@ce.vjti.ac.in}, \\\texttt{sakhedkar@ce.vjti.ac.in}
}

\begin{document}
\maketitle
\begin{abstract}
We present BhashaSetu, a linguistically enriched English--Marathi parallel dataset addressing persistent data limitations in low-resource neural machine translation (NMT). Marathi, spoken by over 95 million people, remains underrepresented in high-quality parallel corpora across diverse domains. Our dataset comprises 2.78 million sentence pairs from heterogeneous sources including news, politics, healthcare, literature, and culture, with stemmed and lemmatized representations to support morphology-aware analysis. We benchmark multiple state-of-the-art translation models using BLEU, spBLEU, chrF++, and TER metrics, and conduct parameter-efficient fine-tuning of NLLB-200-distilled-600M using LoRA. A key finding from our ablation: corpus-level deduplication is the single largest preprocessing contributor to downstream quality (removing it reduces performance by 1.17 BLEU and 2.21 chrF++), demonstrating that disciplined cross-source corpus hygiene is a low-cost, high-impact intervention for low-resource, morphologically rich languages. The dataset is publicly released to promote reproducible and linguistically informed low-resource NMT research.
\end{abstract}

\section{Introduction}
Despite rapid advances in large-scale multilingual models, translation quality for low-resource languages remains substantially behind high-resource counterparts. Model scale alone is insufficient: recent evaluations show that models such as NLLB \citep{nllb2022}, LLaMA 3 \citep{grattafiori2024llama3herdmodels}, and Gemma 2 \citep{gemmateam2024gemma2improvingopen} continue to perform poorly on Indic languages \citep{vaidya2025analysisindiclanguagecapabilities}, indicating that task alignment, linguistic diversity, and data quality are the primary factors driving robust translation.
 
Marathi exemplifies this gap despite being spoken by over 95 million people. High-quality English--Marathi parallel corpora remain scarce, dispersed, and frequently restricted to narrow domains \citep{Siripragrada2020AMPB,Haddow2020PMIndiaAD}. Current resources either provide high-quality alignments at insufficient volume or prioritize scale without systematic curation, causing NMT systems to struggle with agreement errors, incorrect inflections, and semantic drift.
 
English--Marathi translation is further complicated by Marathi's rich inflectional system (tense, aspect, mood, gender, number, and case) and flexible word order \citep{Kowtal2024ADSA,Ojha2021FindingsOTC}. Motivated by a data-centric perspective \citep{SINGH2023144}, we introduce \textbf{BhashaSetu}, a linguistically enriched English--Marathi parallel corpus of 2.78 million sentence pairs across five domains, built with a language-aware preprocessing framework combining systematic cleaning, Unicode normalization, and morphology-aware transformations.
 
Corpus-level deduplication emerges from our controlled ablation as the single most influential preprocessing step: removing it reduces performance by 1.17 BLEU and 2.21 chrF++ (Table~\ref{tab:ablation_results}). Our zero-shot benchmarking of open-source and proprietary models reveals substantial variability and confirms that Marathi has not been a core optimization target in mainstream systems. Our contributions are:
 
\begin{itemize}
\item A domain-diverse English--Marathi parallel corpus of 2.78 million sentence pairs.
\item A generalized, language-aware preprocessing framework for morphologically rich languages.
\item Empirical evidence that dataset quality and linguistic alignment can outweigh model scale in low-resource translation.
\item A per-source license audit with a legally redistributable subset and provenance documentation.
\end{itemize}

\section{Related Work} 
Early English--Marathi resources were small or narrowly focused. The ILCI corpus provides manually curated pairs for several Indian languages but is restricted in scale and domain.\footnote{\url{https://sanskrit.jnu.ac.in/projects/ilci.jsp?proj=ilci}} PMIndia offers up to 56K sentence pairs per language, concentrated in governmental text \citep{Haddow2020PMIndiaAD}. \citet{Siripragrada2020AMPB} compile a 407K-sentence corpus from PIB and Mann Ki Baat, while Anuvaad released approximately 2.54M English--Marathi pairs across legal and news domains \citep{anuvaadcorpus2021}. Samanantar substantially increases coverage through large-scale web mining \citep{samanantar}, and BPCC aggregates roughly 230M sentence pairs across all 22 scheduled Indian languages as part of IndicTrans2 \citep{gala2023indictrans2highqualityaccessiblemachine}. FLORES-200 provides a high-quality evaluation benchmark but is too small for training \citep{flores200}.

On the system side, \citet{mujadia-sharma-2021-english} and \citet{jain-etal-2021-evaluating} demonstrated the value of morphology-aware segmentation and back-translation for English--Marathi NMT; \citet{dabre-etal-2022-indicbart} showed that language-group pretraining can compete with larger generic models. More recent model-side advances include Sarvam-Translate and TranslateGemma \citep{finkelstein2026translategemmatechnicalreport}, while IndicGenBench \citep{singh2024indicgenbench} and WMT24++ \citep{deutsch-etal-2025-wmt24} expand multilingual evaluation coverage. These advances underscore the continuing need for high-quality, linguistically curated parallel corpora to provide reliable supervision and domain diversity for Marathi's morphological and register challenges.
 
Table~\ref{tab:dataset_comparison} compares major English--Marathi resources, highlighting gaps in scale, curation, and domain diversity that motivate BhashaSetu.

\FloatBarrier
\begin{table}[!ht]
\centering
\footnotesize
\setlength{\tabcolsep}{3pt}
\renewcommand{\arraystretch}{1.15}
\resizebox{\columnwidth}{!}{%
\begin{tabular}{p{1.95cm} c c p{2.3cm} p{2.3cm}}
\toprule
\textbf{Dataset} & \textbf{Year} & \textbf{No. of rows (approx)} & \textbf{Strength} & \textbf{Limitation} \\
\midrule
ILCI & 2020 & 50K & Manual alignment & Small scale \\
PMIndia & 2020 & 56K & Public multilingual govt/news corpus & Narrow domain \\
PIB/MKB corpus & 2020 & 407K & 10-language aligned public corpus & Public-information skew \\
Anuvaad & 2021 & 2.54M & Large public English--Marathi release & Govt/legal source skew \\
ULCA/Bhashini & 2022 & Ongoing & Open national repository model & Not a fixed curated corpus \\
Samanantar & 2021 & 3.63M & Large web corpus & Quality variance \\
FLORES-200 & 2022 & 3K & Eval benchmark & Not for training \\
BPCC & 2023 & 230M total & Massive multilingual scale & Limited Marathi-specific curation \\
Sangraha & 2024 & 5.75M & Massive synthetic data & Synthetic shift \\
BhashaSetu (Ours) & 2026 & 2.78M & Domain-diverse, linguistically processed & Formal-text bias \\
\bottomrule
\end{tabular}}
\caption{Comparison of major English--Marathi resources and related multilingual corpora.}
\label{tab:dataset_comparison}
\end{table}

\section{Dataset Construction and Statistics}
\begin{figure*}[t]
\centering
\begin{tikzpicture}[scale=0.9, transform shape, node distance=0.6cm, auto, every node/.style={font=\scriptsize}]
  \node (collect) [startstop, text width=2.6cm, minimum height=1.0cm, fill=gray!20] {\textbf{Collect matching datasets} \\
Find Eng--Mar sentences with the same meaning};
  \node (pick) [process, right=of collect, text width=2.6cm, minimum height=1.0cm] {\textbf{Select useful data} \\
Keep only English and Marathi sentence columns};
  \node (organize) [processgreen, right=of pick] {\textbf{Organize data} \\
Structure the data properly.};
  \node (clean) [processorange, right=of organize] {\textbf{Clean data} \\
Remove duplicates and empty rows};
  \node (consist) [processyellow, right=of clean] {\textbf{Standardize text} \\
Fix format and remove very short/long sentences};
  \node (prepare) [processblue, right=of consist] {\textbf{Prepare text} \\
Convert words to base form (lemmatization/stemming)};

  \draw [arrow] (collect) -- (pick);
  \draw [arrow] (pick) -- (organize);
  \draw [arrow] (organize) -- (clean);
  \draw [arrow] (clean) -- (consist);
  \draw [arrow] (consist) -- (prepare);
\end{tikzpicture}
\caption{Overview of the BhashaSetu preprocessing pipeline, showing how raw bilingual sources are collected, filtered, standardized, and morphologically processed before inclusion in the final English--Marathi translation corpus.}
\label{fig:preprocess_pipeline}
\end{figure*}
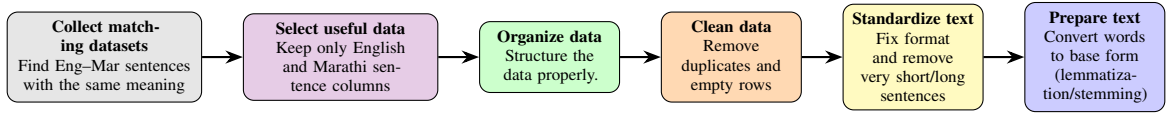

\begin{figure}[t]
\centering
\resizebox{\linewidth}{!}{
\begin{tikzpicture}[scale=0.7, transform shape, node distance=1.0cm, auto, every node/.style={font=\normalsize}]
  \node (start) [startstop, text width=2.2cm, minimum height=0.9cm] {Start};
  \node (batch) [process, below=of start, text width=3.2cm, minimum height=0.9cm] {Divide dataset into small batches};
  \node (select) [process, below=of batch, text width=3.2cm, minimum height=0.9cm] {Select one batch};

\node (eng) [process, below=of select, xshift=-3.2cm, yshift=-1.2cm, text width=3.6cm, minimum height=1.0cm] {\textbf{English processing:} normalize text, tokenize, lemmatize, stemming};
  \node (mar) [process, below=of select, xshift=3.2cm, yshift=-1.2cm, text width=3.6cm, minimum height=1.0cm] {\textbf{Marathi processing:} clean/standardize, tokenize, stemming};

  \node (merge) [process, below=of select, yshift=-3.5cm, text width=3.0cm, minimum height=0.9cm] {Merge cleaned English and Marathi results};
  \node (save) [process, below=of merge, text width=3.0cm, minimum height=0.9cm] {Save processed batch (CSV)};

  \node (check) [process, below=of save, text width=3.0cm, minimum height=0.9cm] {More batches to process?};
  \node (final) [startstop, below=of check, text width=3.2cm, minimum height=0.9cm] {Save final cleaned dataset};
  \node (end)[startstop, below=of final, text width=3.2cm, minimum height=0.9cm] {End};
\draw [arrow] (start) -- (batch);
\draw [arrow] (batch) -- (select);

\draw [arrow] (select.south) -- ++(0,-0.8) -| (eng.north);
\draw [arrow] (select.south) -- ++(0,-0.8) -| (mar.north);

\draw [arrow] (eng.south) |- (merge.west);
\draw [arrow] (mar.south) |- (merge.east);

\draw [arrow] (merge) -- (save);
\draw [arrow] (save) -- (check);

\draw [arrow] (check.west) -- ++(-4,0) node[midway, above]{Yes} |- (select.west);
\draw [arrow] (check.south) -- node[right]{No} (final);

\draw [arrow] (final) -- (end);
\end{tikzpicture}
}
\hfill

\caption{\textbf{Batch-based text processing workflow} \\ Batch execution enables scalable preprocessing while preserving language-specific transformations.}
\label{fig:batch_pipeline}
\end{figure}
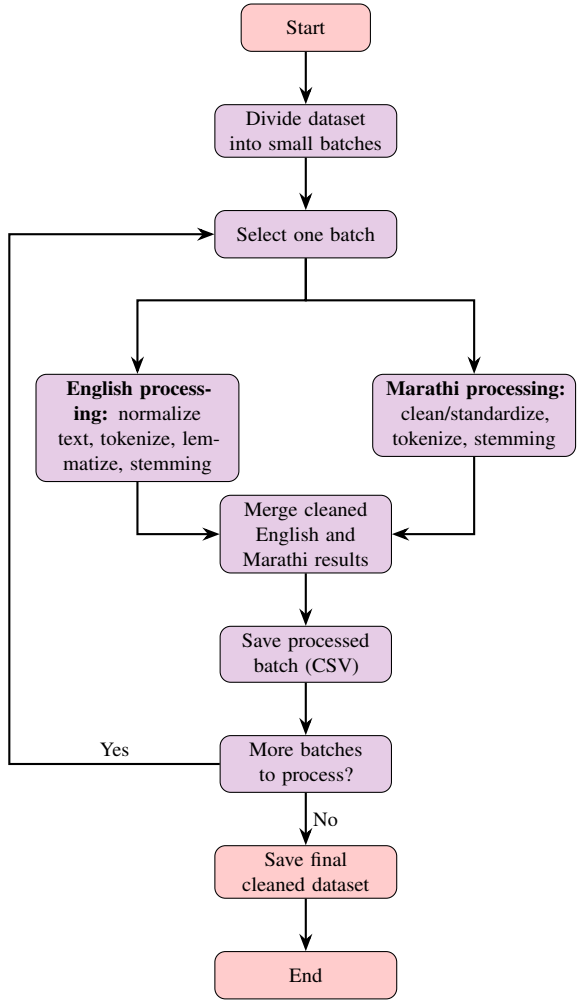

\begin{figure*}[t]
\centering
\includegraphics[width=\textwidth]{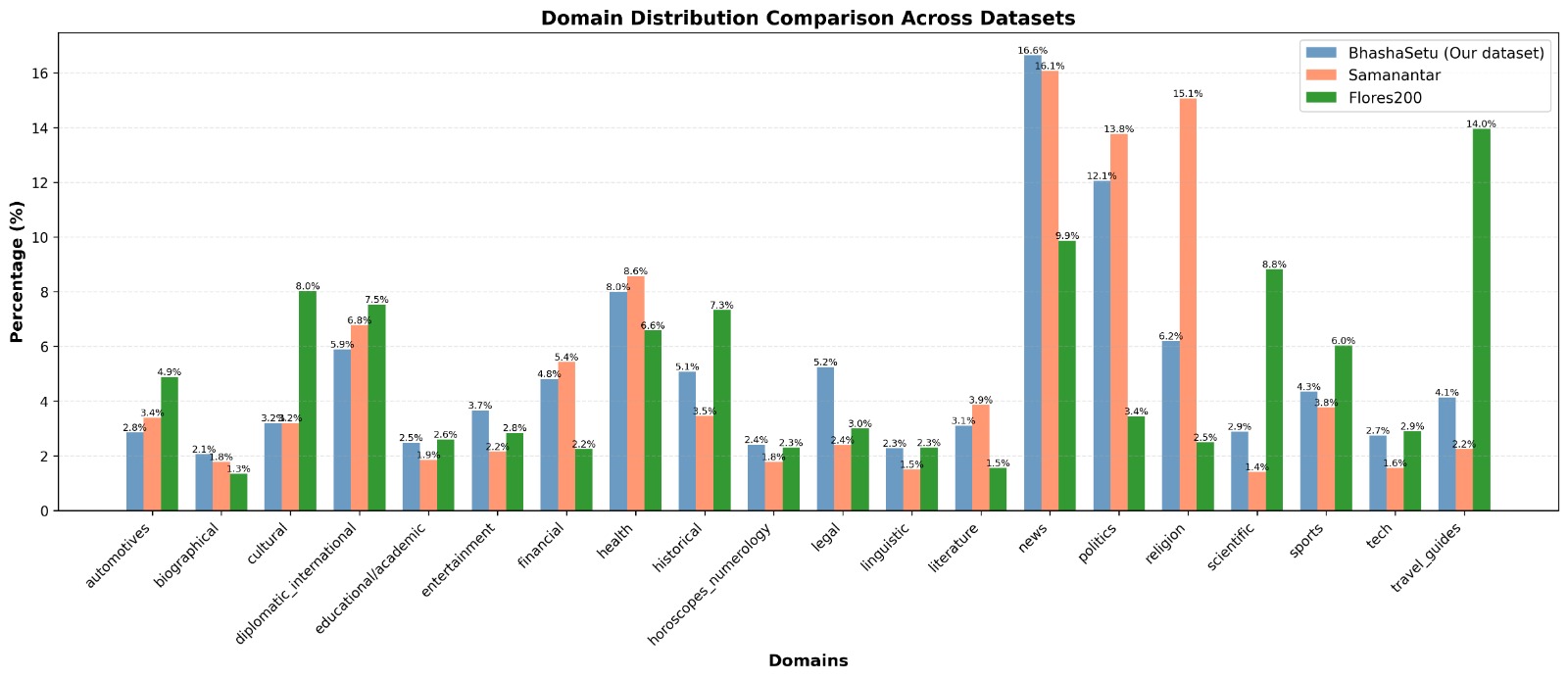}
\caption{Domain distribution of the dataset. The corpus spans multiple domains, with the largest share concentrated in news, religion, and politics, reflecting source availability while still preserving topical diversity for translation evaluation and training.}
\label{fig:dataset_domain}
\end{figure*}

Dataset construction follows three principles: linguistic consistency, domain diversity, and reproducible preprocessing. Figure~\ref{fig:preprocess_pipeline} summarizes the pipeline.
 
We collect existing English--Marathi parallel datasets, extract paired sentence columns, and merge them into a raw CSV. Cleaning steps include: deduplication on both columns, null/empty row removal, and Unicode NFC normalization for Marathi's Devanagari script. Case information is retained on the English side to preserve named-entity cues. NLP tools (spaCy, NLTK for English; Indic NLP Library for Marathi) are initialized once. We apply a sentence length filter of 3--50 words per sentence as a quality-control heuristic: very short pairs were disproportionately associated with headers and fragments, while very long pairs more often reflected sentence-boundary errors or paragraph-level misalignments. After all preprocessing, the final corpus contains 2,779,901 sentence pairs, from which 50,000 are held out as a stratified evaluation subset.
 
Table~\ref{tab:preprocessing_retention} shows that deduplication and length filtering account for nearly all row removal; linguistic normalization steps are lossless.

\begin{table}[t]
\centering
\footnotesize
\setlength{\tabcolsep}{4pt}
\renewcommand{\arraystretch}{1.1}
\resizebox{\columnwidth}{!}{%
\begin{tabular}{l r r r r}
\toprule
\textbf{Preprocessing Stage} & \textbf{Before (\%)} & \textbf{After (\%)} & \textbf{Removed (\%)} & \textbf{Retained (\%)} \\
\midrule
Duplicate Removal        & 100.00 &  99.99 & 0.01 &  99.99 \\
Null Row Removal         & 100.00 & 100.00 & 0.00 & 100.00 \\
Unicode Normalization    & 100.00 & 100.00 & 0.00 & 100.00 \\
Length Filtering         & 100.00 &  99.67 & 0.33 &  99.67 \\
English Tokenization     & 100.00 & 100.00 & 0.00 & 100.00 \\
English Lemmatization    & 100.00 & 100.00 & 0.00 & 100.00 \\
English Stemming         & 100.00 & 100.00 & 0.00 & 100.00 \\
Marathi Tokenization     & 100.00 & 100.00 & 0.00 & 100.00 \\
Marathi stemming & 100.00 & 100.00 & 0.00 & 100.00 \\
\bottomrule
\end{tabular}%
}
\caption{Per-stage percentage retention of the preprocessing pipeline (Before = 100\%). Deduplication and length filtering account for most loss; cumulative retention $\approx 99.66\%$.}
\label{tab:preprocessing_retention}
\end{table}
We acknowledge that this filtering threshold introduces a design limitation. Legal texts, literary prose, and technical documentation frequently contain sentences exceeding 50 words that are syntactically well-formed and translationally valid. Similarly, certain formulaic or conversational genres produce legitimate sentences below 3 words. By excluding these, the corpus may underrepresent stylistic and domain-specific variation that is important for building robust MT systems. Future iterations could adopt adaptive, domain-conditioned length thresholds or length-stratified quality scoring to retain a broader range of valid sentence structures while still suppressing alignment noise. Finally, the processed texts are merged, and the dataset is saved, yielding approximately 7GB of data with millions of parallel sentences suitable for NMT. After all preprocessing stages, the final released corpus contains 2,779,901 sentence pairs, from which 50,000 are held out as a stratified evaluation subset (see Section 4) and the remaining 2,729,901 form the training split.

As a coarse semantic check, we computed LaBSE cosine similarity over all sentence pairs (mean 0.7086, std 0.1082, min 0.1491, max 0.9688). Approximately 3.1\% of pairs fall below 0.5 and 8.7\% below 0.6. We retain these pairs because LaBSE penalizes domain-specific and named-entity-rich pairs where surface-form divergence is linguistically valid, and because a pilot LaBSE-filtered run ($\Delta\text{BLEU} < 0.1$) showed no material benefit from filtering.

\begin{table}[t]
\centering
\footnotesize
\setlength{\tabcolsep}{5pt}
\renewcommand{\arraystretch}{1.1}
\begin{tabular}{lcccc}
\toprule
\textbf{Dataset} & \textbf{Mean} & \textbf{Std} & \textbf{Min} & \textbf{Max} \\
\midrule
BhashaSetu (Ours) & 0.7086 & 0.1082 & 0.1491 & 0.9688 \\
Samanantar & 0.7886 & 0.1471 & -0.2695 & 0.9929 \\
FLORES-200 & 0.8578 & 0.0464 & 0.4944 & 0.9604 \\
\bottomrule
\end{tabular}
\caption{Comparison of LaBSE cosine similarity statistics across BhashaSetu, Samanantar, and FLORES-200.}
\label{tab:labse_comparison}
\end{table}

Figure~\ref{fig:dataset_domain} shows that the corpus spans multiple domains but is concentrated in News (16.6\%), Religion (15.1\%), and Politics (13.0\%), reflecting source availability while preserving broader topical diversity. This mixture of formal news, religious/cultural, and political text enables more realistic in-domain adaptation and evaluation than smaller, more narrowly focused corpora: models can be validated on heterogeneous registers and domain-stratified held-out splits rather than a single narrow distribution. Compared with datasets that are heavily web-mined or benchmark-limited, BhashaSetu's deliberate domain variety supports cross-domain fine-tuning experiments, robustness checks for register sensitivity, and targeted collection of underrepresented genres in follow-up releases.

\subsection{Source Composition}
BhashaSetu is assembled from existing publicly available English--Marathi corpora; no new alignments were created. Table~\ref{tab:source_breakdown} provides a per-source breakdown. The novel contribution lies in the unified curation pipeline: cross-source deduplication, provenance-aware merging, language-specific normalization for Devanagari, license-aware source selection, and domain-stratified held-out splits. These coordinated choices change corpus composition in ways that matter for downstream training, as confirmed by the ablation study (Table~\ref{tab:ablation_results}) and fine-tuning comparisons (Table~\ref{tab:nllb_finetuning_scores}).

\begin{table}[t]
\centering
\footnotesize
\setlength{\tabcolsep}{3pt}
\renewcommand{\arraystretch}{1.15}
\resizebox{\columnwidth}{!}{%
\begin{tabular}{l r p{1.6cm} c l c}
\toprule
\textbf{Source} & \textbf{Rows (K)} & \textbf{Domain(s)} & \textbf{Period} & \textbf{License} & \textbf{New?} \\
\midrule
Anuvaad          & 1,784 & Legal, news, general      & 2018--2021 & CC BY 4.0       & No \\
BPCC (en--mr)    & 288 & Mixed multilingual         & 2020--2023 & CC0 / CC BY 4.0 & No \\
Samanantar (en--mr)  & 677  & news, politics, mixed   & 2021 & CC BY-NC 4.0 & No \\
aiKosh           &  2 & sports, law, mixed & 2025      & CC BY-SA 4.0 & Yes \\
PMIndia          &    4 & Government, news           & 2019--2020 & Open            & No \\
FLORES-200       &     3 & news, sports             & 2022             & CC BY-SA 4.0  & No \\
Other public     &   31 & Religion, culture, health  & Various    & Various open    & No \\
\midrule
\textbf{Total (raw)}   & \textbf{2,789} \\
\textbf{Total (final)} & \textbf{2,780} \\
\bottomrule
\end{tabular}}
\caption{Source breakdown of BhashaSetu after preprocessing. ``New?'' indicates whether the source was newly ingested for this release.}
\label{tab:source_breakdown}
\end{table}

\section{Benchmark Experiments}

\paragraph{Computational Infrastructure.}
Zero-shot benchmarking of smaller models (opus-mt-en-mr, Misal-1B, LLaMA-3.2-1B) was conducted on an NVIDIA RTX 5090 (12\,GB VRAM), while larger models (Tiny-Aya-Global, IndicTrans2-1B) and all 
fine-tuning experiments were run on an NVIDIA DGX A100. 

We evaluate a mix of open-source and closed-source models for English-to-Marathi translation in two phases. First, zero-shot evaluation: models translate held-out sentences scored by BLEU, chrF++, and TER \citep{Papineni2002BleuAM,popovic-2017-chrfpp,snover-etal-2006-study}. Higher BLEU and chrF++ indicate better quality; higher TER indicates worse quality.
 
Evaluation uses 50,000 held-out sentence pairs from BhashaSetu, 50,000 from Samanantar (both stratified by domain), and the full FLORES-200 English--Marathi benchmark (3,001 sentences). Fine-tuning uses the remaining 2.73M BhashaSetu training pairs. Key inference hyperparameters: beam size 4, deterministic decoding, max generation 128 tokens, forced Marathi BOS for NLLB.

\begin{table*}[!ht]
\centering
\footnotesize
\setlength{\tabcolsep}{3.5pt}
\resizebox{\textwidth}{!}{%
\begin{tabular}{lcccccccccccc}
\toprule
 & \multicolumn{4}{c}{\textbf{BhashaSetu (Ours)}} 
 & \multicolumn{4}{c}{\textbf{Samanantar}} 
 & \multicolumn{4}{c}{\textbf{Flores200}} \\
\cmidrule(lr){2-5} \cmidrule(lr){6-9} \cmidrule(lr){10-13}
\textbf{Model} 
& \textbf{BLEU} & \textbf{spBLEU} & \textbf{chrF++} & \textbf{TER}
& \textbf{BLEU} & \textbf{spBLEU} & \textbf{chrF++} & \textbf{TER}
& \textbf{BLEU} & \textbf{spBLEU} & \textbf{chrF++} & \textbf{TER} \\
\midrule
opus-mt-en-mr   & 0.03 & 0.02 & 0.04 & 400.34 & 1.1681 & 6.2343 & 11.5525 & 131.1696 & 0.1568 & 2.0663 & 10.4031 & 108.7911 \\
Misal-1B        & 0.01 & 0.01 & 0.12 & 110.62 & 0.2362 & 2.5864 & 1.0563 & 141.2023 & 0.0921 & 0.0874 & 0.6059 & 222.5148 \\
LLaMA-3.2-1B    & 0.04 & 0.04 & 1.21 & 142.04 & 0.1468 & 1.0504 & 3.7155 & 339.1645 & 0.3161 & 1.2516 & 3.6652 & 210.4155 \\
Tiny-Aya-Global       & 1.2570 & 1.5848 & 15.5772 & 109.3641 & 1.3930 & 2.2467 & 21.5171 & 324.1271 & 2.6594 & 6.6582 & 30.2313 & 448.8505 \\
IndicTrans2-1B  & 7.0408 & 7.1144 & 43.4382 & 88.0192 & 12.62 & 12.81 & 42.50 & 84.33 & 0.0009 & 0.4882 & 2.6689 & 98.1948 \\
\bottomrule
\end{tabular}
}

\caption{Translation performance of different models across three English--Marathi datasets using BLEU, spBLEU, chrF++, and TER metrics.}
\label{tab:mt_scores}
\end{table*}

\vspace{3pt}
\noindent\textbf{Note on extreme TER / anomalous rows:} Several unusually large TER values (well above the typical 0--200 range) reflect decoding or format failure modes (for example, empty or truncated outputs, script/encoding mismatches, or severe tokenization/detokenization differences) rather than interpretable translation quality. Spot checks indicate the following likely causes by row: \textit{opus-mt-en-mr} on BhashaSetu (TER=400.34) produced many short/blank or script-mismatched outputs; \textit{Misal-1B} on FLORES-200 (TER=222.51) often generated extremely short outputs; \textit{LLaMA-3.2-1B} on Samanantar (TER=339.16) and FLORES-200 (TER=210.42) showed truncation/format sensitivity; \textit{Tiny-Aya-Global} on Samanantar (TER=324.13) and FLORES-200 (TER=448.85) exhibited tokenization/length-mismatch failures. Finally, \textit{IndicTrans2-1B} on FLORES-200 (BLEU=0.0009, chrF++=2.67) appears consistent with a tokenization/detokenization mismatch (very low token overlap but some character overlap). These rows should be interpreted as reporting artifacts from failed decodings rather than reliable indicators of model quality.

\begin{table*}[!ht]
\centering
\footnotesize
\setlength{\tabcolsep}{3.5pt}
\resizebox{\textwidth}{!}{%
\begin{tabular}{lcccccccccccc}
\toprule
 & \multicolumn{4}{c}{\textbf{BhashaSetu (Ours)}}
 & \multicolumn{4}{c}{\textbf{Samanantar}}
 & \multicolumn{4}{c}{\textbf{Flores200}} \\
\cmidrule(lr){2-5} \cmidrule(lr){6-9} \cmidrule(lr){10-13}
\textbf{Dataset}
& \textbf{BLEU} & \textbf{spBLEU} & \textbf{chrF++} & \textbf{TER}
& \textbf{BLEU} & \textbf{spBLEU} & \textbf{chrF++} & \textbf{TER}
& \textbf{BLEU} & \textbf{spBLEU} & \textbf{chrF++} & \textbf{TER} \\
\midrule
NLLB + BhashaSetu finetuning   & 9.8639 & 20.5969 & 45.2385 & 80.2816 & 7.5042 & 17.0216 & 39.5152 & 90.1038 & 10.4393 & 23.4542 & 43.8343 & 76.3734 \\
NLLB + Flores200 finetuning    & 5.5321 & 14.1363 & 41.1097 & 89.3282 & 4.7412 & 11.7273 & 34.3082 & 88.6872 & 17.0623 & 28.6089 & 51.8482 & 69.5062 \\
NLLB + Samanantar finetuning  & 8.5107 & 16.7119 & 38.4339 & 89.1385 & 10.2413 & 18.1504 & 39.2149 & 84.8700 & 12.9023 & 23.7962 & 46.7300 & 75.0890 \\
\bottomrule
\end{tabular}
}
\caption{Translation performance of NLLB LoRA+PEFT fine-tuning variants (all parameters updated) across three English--Marathi datasets using BLEU, spBLEU, chrF++, and TER metrics.}
\label{tab:nllb_finetuning_scores}
\end{table*}

We summarize the main inference hyperparameters used for decoding in Table~\ref{tab:inference_settings}.
\begin{table}[t]
\centering
\footnotesize
\begin{tabular}{p{2.2cm} p{4.8cm}}
\toprule
\textbf{Parameter} & \textbf{Value} \\
\midrule
Beam size & 4 (Indic flows often use 5) \\
Decoding & Beam search (greedy for non-Indic/chat paths) \\
Sampling & Deterministic (do\_sample=false) \\
Max generation & 64--128 tokens (benchmark default: 128) \\
Notes & Forced Marathi BOS for NLLB; early\_stopping=true \\
\bottomrule
\end{tabular}
\caption{Key inference hyperparameters }
\label{tab:inference_settings}
\end{table}

\section{Parameter-Efficient Fine-Tuning Setup}
\label{sec:param_efficient_setup}
We fine-tuned NLLB-200-distilled-600M \citep{nllb2022} on the English--Marathi training split of BhashaSetu using Low-Rank Adaptation (LoRA) \citep{hu2022lora}. LoRA offers a parameter-efficient alternative to full-model updating and is well suited to low-resource adaptation. We selected LoRA for several practical reasons: it trains orders of magnitude fewer parameters than full fine-tuning (reducing GPU memory and checkpoint size), tends to converge faster in low-data regimes, is easy to share and deploy as a small delta on top of a public backbone, and empirically matches or approaches full fine-tuning performance for many sequence-to-sequence tasks. Compared with adapter modules or prefix-tuning, LoRA provides a simpler integration path for transformers used in NMT backbones and keeps the runtime graph unchanged while enabling efficient backpropagation through low-rank updates. These operational and empirical trade-offs made LoRA a suitable choice for reproducible, parameter-efficient adaptation in our experiments. We used the NLLB language tags \texttt{eng\_Latn} and \texttt{mar\_Deva}, set the forced decoder beginning-of-sequence (BOS) token to Marathi, and trained on the Hugging Face release in streaming mode.

Training used a maximum sequence length of 128, per-device batch size 8, gradient accumulation 4, learning rate \(1.5 \times 10^{-4}\), 1,500 warmup steps, and 30,000 total steps. We enabled bfloat16 on supported graphics processing units (GPUs) and applied label smoothing of 0.1. For LoRA, we used \(r=16\), \(\alpha=32\), and dropout 0.05 over \texttt{q\_proj}, \texttt{k\_proj}, \texttt{v\_proj}, and \texttt{out\_proj}. These values were chosen as a standard stable low-rank configuration following common LoRA practice, with \(\alpha=32\) preserving the usual \(\alpha/r=2\) scaling and dropout 0.05 providing mild regularization \citep{hu2022lora}. We did not perform a separate hyperparameter ablation, so this setting should be viewed as a reproducible default rather than a tuned optimum. Batches were dynamically padded with \texttt{DataCollatorForSeq2Seq}.

\begin{figure}[t]
\centering
\includegraphics[width=\linewidth]{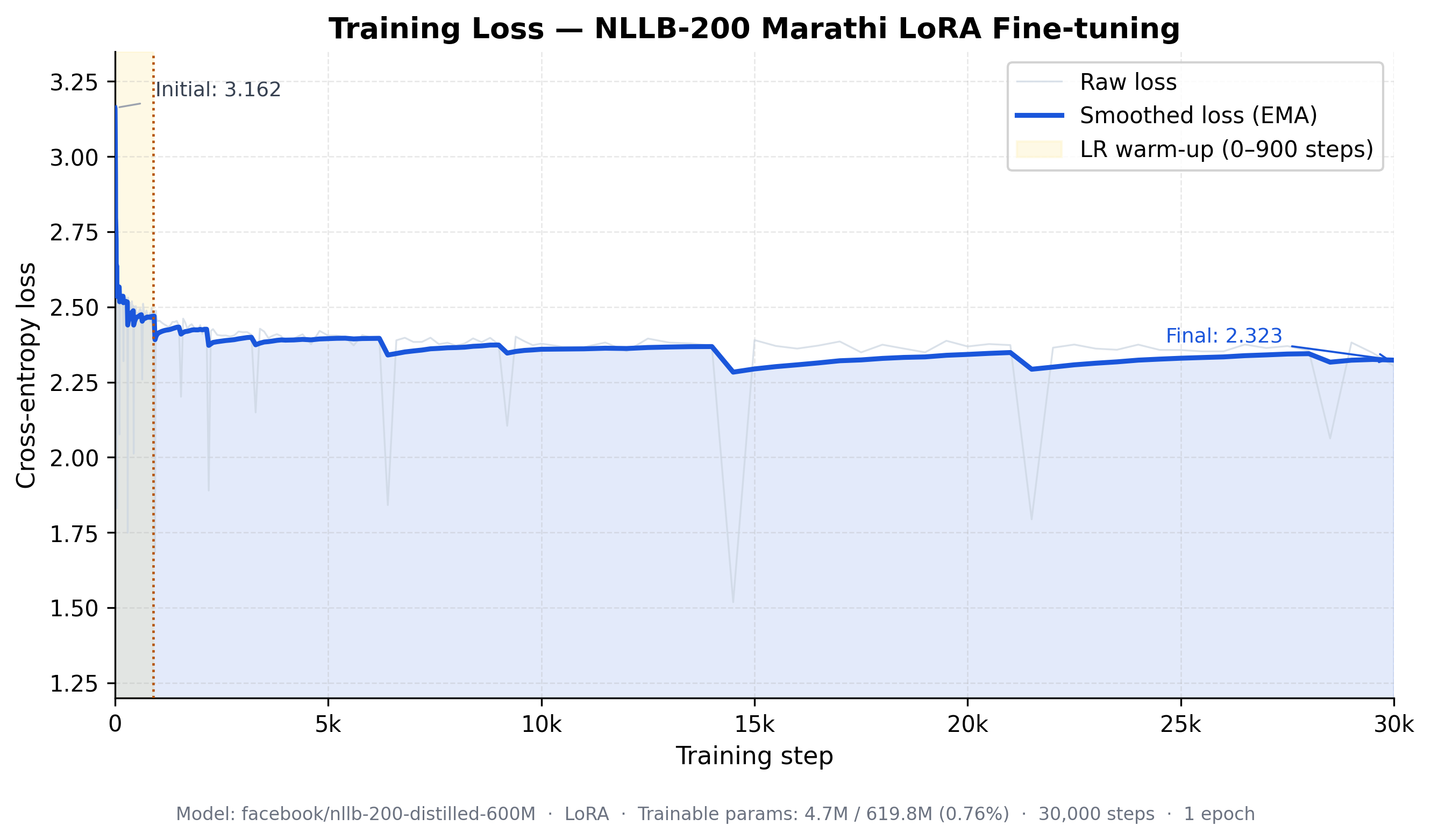}
\caption{Training curves for LoRA-based fine-tuning of NLLB-200-distilled-600M on the English--Marathi training split of BhashaSetu. The plot tracks optimization behavior across training steps and is used to monitor stability during parameter-efficient adaptation.}
\label{fig:nllb_lora_training_curves}
\end{figure}

These experiments are included to assess whether BhashaSetu supports stable and effective parameter-efficient adaptation of a strong multilingual NMT backbone. Quantitative outcomes are discussed with the main experimental results, while Figure~\ref{fig:nllb_lora_training_curves} summarizes the training dynamics observed during fine-tuning.

\section{Results and Analysis}
We summarize the evaluation metrics briefly and then discuss model behavior across datasets and preprocessing settings.

\subsection{Evaluation Metrics}
We report BLEU \citep{Papineni2002BleuAM}, sentencepiece BLEU (spBLEU) \citep{flores200,kudo-richardson-2018-sentencepiece}, chrF++ \citep{popovic-2017-chrfpp}, TER \citep{snover-etal-2006-study}, and COMET \citep{rei-etal-2020-comet}. chrF++ combines character n-gram precision/recall with word-level unigram signals and is therefore particularly informative for morphologically rich Marathi while also capturing word-level adequacy. BLEU and spBLEU capture token- and subword-level overlap, chrF++ complements them by emphasizing character-level morphology with a unigram signal, and TER measures post-edit distance (lower is better). COMET is a learned neural metric trained to correlate with human judgements and is useful for capturing adequacy and fluency signals that overlap-based metrics can miss; in a few visualizations (e.g., the ablation figure) we report scaled COMET differences as \(\Delta\text{COMET}\times 100\) for readability. We therefore report these metrics jointly rather than relying on any single score. For several unusually weak systems, a small manual diagnostic suggested likely issues such as script mismatches, prompt-format sensitivity, and occasional decoding truncation, which may also explain some large BLEU--spBLEU gaps. 

\noindent\textit{For discussion of tokenization and normalization effects on BLEU and recommendations for reproducible scoring, see} \citet{post-2018-call} \textit{and the sacreBLEU documentation; note that FLORES uses a SentencePiece preprocessing regime and that character-level metrics such as chrF++ are comparatively more robust to script/tokenization mismatches} \citep{popovic-2017-chrfpp,flores200,husain2024romansetu}.

\subsection{Analysis of Zero Shot Benchmarking results}
Table~\ref{tab:mt_scores} illustrates the ongoing challenge of English-to-Marathi translation in zero-shot settings. The majority of evaluated models, including large multilingual and instruction-tuned LLMs, consistently have low BLEU scores, suggesting that Marathi’s rich morphology and flexible word order cannot be handled by model scale alone. This observation is consistent with recent analyses documenting systematic weaknesses of general-purpose LLMs on Indic languages, particularly in low-resource and morphologically complex scenarios \citep{vaidya2025analysisindiclanguagecapabilities}. These results underscore the importance of language-specific data and inductive biases for effective translation into Indian languages.

Among the evaluated systems, IndicTrans2-1B is the strongest model on BhashaSetu and Samanantar by a clear margin, particularly on BLEU and chrF++, indicating that language-specific multilingual NMT pretraining remains highly valuable for English--Marathi translation. In contrast, smaller general-purpose models such as Tiny-Aya-Global and LLaMA-3.2-1B show limited zero-shot performance, with modest chrF++ but weak BLEU and high TER, suggesting partial lexical or semantic overlap without reliable structural fidelity. The weakest results are observed for \textit{opus-mt-en-mr} and Misal-1B, whose low overlap-based scores and high edit distance indicate substantial translation breakdowns under our evaluation setup. Overall, the divergence between chrF++ and TER across models highlights the need for complementary evaluation metrics and validates our dataset as a challenging benchmark for English--Marathi NMT.

\subsection{Analysis of Fine-Tuning Results}
Table~\ref{tab:nllb_finetuning_scores} (full-model NLLB fine-tuning; all model parameters updated) shows that dataset choice strongly affects multilingual NMT adaptation. Fine-tuning on BhashaSetu yields the best performance on the BhashaSetu test set, reaching 9.86 BLEU, 20.60 spBLEU, 45.24 chrF++, and 80.28 TER, which indicates that its curated and linguistically processed supervision provides a strong adaptation signal for English--Marathi translation. Compared with FLORES-200 fine-tuning, BhashaSetu fine-tuning is also markedly more robust as a training resource: although FLORES-200 achieves the strongest scores on its own benchmark, its much smaller size leads to substantially weaker transfer to BhashaSetu and Samanantar. Samanantar fine-tuning remains competitive, especially on Samanantar and FLORES-200, but the stronger in-domain gains from BhashaSetu support our broader claim that cleaner, domain-diverse, and morphology-aware corpora can be more effective for multilingual NMT adaptation than benchmark-oriented or noisier large-scale alternatives.

Clarification: Table~\ref{tab:nllb_finetuning_scores} reports full-model fine-tuning runs (updating all parameters), while Table~\ref{tab:ablation_results} reports parameter-efficient LoRA adaptations on the distilled 600M NLLB backbone (see Section~\ref{sec:param_efficient_setup}). These regimes differ in trainable parameters and optimization dynamics, which explains the lower absolute scores in the ablation table compared with the full-model fine-tuning numbers.

\section{Ablation Studies}
To quantify the contribution of each preprocessing component, we conduct an ablation study using NLLB-600M LoRA on the BhashaSetu test split. The full pipeline (\texttt{C0\_full}) includes lemmatization, stemming, Indic normalization, sentence-length filtering, and deduplication. We remove one component at a time.

\begin{table}[!ht]
\centering
\footnotesize
\setlength{\tabcolsep}{6pt}
\renewcommand{\arraystretch}{1.08}
\begin{tabular}{lcccc}
\toprule
\textbf{Configuration} & \textbf{BLEU} & \textbf{spBLEU} & \textbf{chrF++} & \textbf{TER} \\
\midrule
\texttt{C0\_full}         & \textbf{6.51} & \textbf{15.67} & \textbf{42.02} & \textbf{86.98} \\
\texttt{C1\_no\_lemma}    & 5.98 & 14.61 & 40.79 & 87.15 \\
\texttt{C2\_no\_stem}     & 5.95 & 14.62 & 40.65 & 86.99 \\
\texttt{C3\_no\_indic}    & 5.96 & 14.57 & 40.79 & 87.37 \\
\texttt{C4\_no\_length}   & 5.31 & 14.00 & 39.61 & 87.92 \\
\texttt{C5\_no\_dedup}    & 4.87 & 13.24 & 38.69 & 88.08 \\
\texttt{C6\_no\_linguistic} & 6.04 & 14.67 & 40.90 & 85.98 \\
\bottomrule
\end{tabular}
\caption{Ablation results. Higher BLEU, spBLEU, and chrF++ are better; lower TER is better.}
\label{tab:ablation_results}
\end{table}
 
\begin{figure*}[!tbp]
\centering
\includegraphics[width=\textwidth]{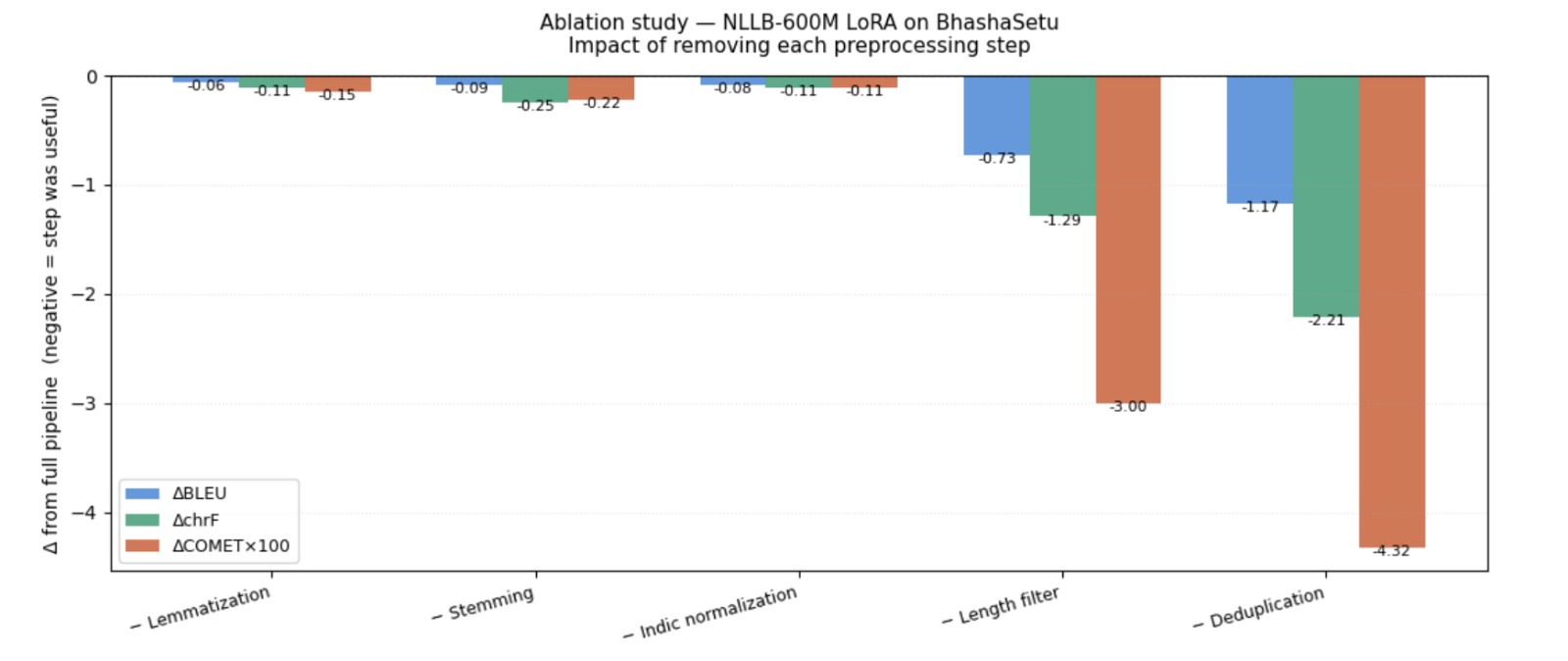}
\caption{Ablation impact relative to the full pipeline. Bars show $\Delta$BLEU, $\Delta$chrF++, and $\Delta$COMET$\times$100; negative values indicate performance drops.}
\label{fig:ablation_image}
\end{figure*}

Removing deduplication (\texttt{C5}) causes the largest drop: 1.17 BLEU and 2.21 chrF++. Length filtering (\texttt{C4}) is the second most influential, reducing BLEU by 0.73 and chrF++ by 1.29. Removing lemmatization, stemming, or Indic normalization individually causes only 0.06--0.09 BLEU degradation, confirming these linguistic steps provide modest but consistent improvements over a subword-only baseline. These findings establish corpus-level quality controls deduplication and length filtering as the primary drivers of downstream translation quality, while linguistic normalization contributes smaller additive gains that may be more valuable for morphological error analysis than for automatic metric scores directly.

\section{Dataset Release \& Ethics} 
BhashaSetu will be released under CC BY-SA 4.0, restricted to material whose upstream terms permit redistribution. FLORES-200 (CC BY-SA 4.0), PMIndia (open), Anuvaad (CC BY 4.0), ULCA/Bhashini (CC0 1.0), BPCC (CC0/CC BY 4.0), and Sangraha (CC BY 4.0) are compatible with downstream redistribution. Samanantar (CC BY-NC 4.0) restricts commercial reuse and is excluded from the redistributed text. ILCI requires a separate acquisition process and is also excluded. The public release therefore covers only the filtered subset with explicitly compatible upstream rights.
 
We note two primary limitations. First, approximately 15\% of the Religion category originates from public-domain religious commentaries and cultural websites, introducing topical and lexical bias. Second, the corpus is skewed toward formal-register sources (news, government, legal), so models trained on it may produce overly formal translations and underperform on conversational or dialectal inputs. We document these limitations and planned mitigations (register labelling, targeted conversational data collection) in the dataset README.

In accordance with ACL ARR guidelines, the following responsible research practices are observed:
\begin{itemize}
\item \textbf{Transparency and Reproducibility:} Sources, preprocessing, annotation formats, and predefined splits are fully documented.
\item \textbf{Data Sourcing and Licensing:} Only sources with compatible upstream licenses are redistributed; non-commercial and unclear-license material is excluded.
\item \textbf{Privacy:} No private or restricted data was collected; identifiable details were removed during preprocessing.
\item \textbf{Bias and Representativeness:} The corpus over-represents formal written Marathi; spoken and dialectal varieties are underrepresented.
\item \textbf{Risk and Limitations:} Formal-domain bias may reduce generalization to informal inputs.
\item \textbf{Intended Use:} The dataset is intended for research and model development only, not surveillance or profiling.
\end{itemize}

\section{Use of AI Assistants}
AI assistants were used in the preparation of this paper in the 
following limited capacities: (1) literature search and discovery, 
to identify relevant related work and recent publications on 
low-resource machine translation and Indic NLP; (2) summarization 
of retrieved papers to aid comprehension and comparison; and 
(3) drafting assistance for the abstract, which was subsequently 
reviewed, edited, and verified by the authors. All experimental 
design, data curation, preprocessing, model training, evaluation, 
and analysis were conducted entirely by the authors. The authors 
take full responsibility for the accuracy and integrity of all 
content in this paper.

\section{Conclusions and Future Work} 
BhashaSetu demonstrates that data quality, linguistic preprocessing, and domain diversity remain central to improving low-resource English--Marathi NMT. Beyond providing a 2.78M-sentence parallel corpus, we offer a reproducible framework for building and evaluating morphology-aware NMT resources. Our key finding that corpus-level deduplication is the single largest preprocessing lever provides a broadly applicable, low-cost intervention for other morphologically rich, low-resource language pairs.
 
Future work will extend this pipeline to other low-resource languages, explore Marathi-aligned sentence embeddings, and investigate concept-level multilingual generation with Large Concept Models \citep{qiu2026unified,lcm2024}.

\section{Limitations}
BhashaSetu has several limitations readers should note. The corpus is assembled entirely from existing public corpora, so coverage reflects upstream source choices rather than a purpose-collected sample. Some sources carry non-commercial or unclear licenses, reducing the redistributable subset. The corpus is skewed toward formal written domains; models may underperform on conversational or dialectal inputs. A conservative 3--50 word length filter excludes valid captions and complex clauses. Some weakly aligned pairs remain despite LaBSE screening, and decoding or tokenization failures in a few evaluated systems produce anomalous TER/BLEU values. Automatic metrics are imperfect proxies for human judgment, especially for morphologically rich languages.


\bibliography{custom}

\appendix

\end{document}